\documentclass[10pt,twocolumn,letterpaper]{article}

\usepackage{cvpr}
\usepackage{times}
\usepackage{epsfig}
\usepackage{graphicx}
\usepackage{amsmath}
\usepackage{amssymb}

\usepackage{url}
\usepackage{amsthm, dsfont, accents,amssymb}
\usepackage{subcaption, booktabs}
\usepackage{algorithm}
\usepackage{algorithmic}
\usepackage{multirow,multicol}
\usepackage{enumerate}
\usepackage{mwe}
\usepackage{lipsum}
\usepackage{changes}

\usepackage[flushleft]{threeparttable} 

\usepackage{etoolbox}
\newbool{DrawFigures}
\booltrue{DrawFigures} 

\newcommand{\nn}{\nonumber}
\newcommand{\R}{\mathbb{R}}

\def\ie{{\em i.e.,~}}
\def\eg{{\em e.g.,~}}

\usepackage[pagebackref=true,breaklinks=true,letterpaper=true,colorlinks,bookmarks=false]{hyperref}

\cvprfinalcopy 

\def\cvprPaperID{****} 

\ifcvprfinal\fi
\begin{document}

\title{Deep Unsupervised Clustering Using Mixture of Autoencoders}

\author{Dejiao Zhang\thanks{Part of this work was done when Dejiao Zhang was doing an internship at Technicolor Research. Both Dejiao Zhang and Laura Balzano's participations were funded by DARPA-16-43-D3M-FP-037. Contact information: {dejiao@umich.edu, Yifan.0.sun@gmail.com, eriksson@adobe.com and giralsole@umich.edu.}}\\
University of Michigan\\
Ann Arbor, MI\\
\and
Yifan Sun\\
Technicolor\\
Los Altos, CA\\
\and
Brian Eriksson\\
Adobe\\
San Jose, CA\\
\and
Laura Balzano\\
University of Michigan\\
Ann Arbor, MI\\
}
\maketitle

\begin{abstract}
Unsupervised clustering is one of the most fundamental challenges in machine learning.  A popular hypothesis is that data are generated from a union of low-dimensional nonlinear manifolds; thus an approach to clustering is identifying and separating these manifolds. In this paper, we present a novel approach to solve this problem by using a mixture of autoencoders. 
Our model consists of two parts: 1) a collection of autoencoders where each autoencoder learns the underlying manifold of a group of similar objects, and 2) a mixture assignment neural network, which takes the concatenated latent vectors from the autoencoders as input and infers the distribution over clusters. By jointly optimizing the two parts, we simultaneously assign data to clusters and learn the underlying manifolds of each cluster. 
\end{abstract}

\section{Introduction} 
\label{sec:intro}
The deep learning revolution has been fueled by the explosion of large scale datasets with meaningful labels.  Applications range from image classification~\cite{deng2009imagenet} to text sentiment classification~\cite{pak2010twitter}.  However, for many applications, we do not have meaningful data labels available.
{In this regime, deep autoencoders are gaining momentum \cite{hinton2006reducing} as a way to effectively map data to a low-dimensional feature space where data are more separable and hence more easily clustered \cite{xie2016unsupervised}.}

Long-established methods for unsupervised clustering such as K-means and Gaussian mixture models (GMMs) are still the workhorses of many applications due to their simplicity. However, their distance measures are limited to local relations in the data space and they tend to be ineffective for high dimensional data that often has significant overlaps across clusters \cite{xie2016unsupervised,zheng2016variational}. 

Recently, there has been a surge of interest in developing more powerful clustering methods by leveraging deep neural networks. Various methods \cite{zheng2016variational,dilokthanakul2016deep,zheng2016variational,peng2017deep} have been proposed to conduct clustering on the latent representations learned by (variational) autoencoders. Although these methods perform well in clustering, a weakness is that they use one single low-dimensional manifold to represent the data. As a consequence, for more complex data, the latent representations can be poorly separated. Learning good representations by leveraging the underlying structure of the data has been left largely unexplored {and is the topic of our work. Our underlying assumption is that each data cluster is associated with a separate manifold. 
Therefore, modeling the dataset as a mixture of low-dimensional nonlinear manifolds 
is a natural and promising framework for clustering data generated from different categories.}

{In this paper we develop a novel deep architecture for multiple manifold clustering. Manifold learning and clustering has a rich literature, with parametric estimation methods \cite{elhamifar2011sparse,souvenir2005manifold} and spectral methods being the most common approaches \cite{von2007tutorial,ng2002spectral}. These methods require either a parametric model or distance metrics that capture the relationship among points in the dataset (or both). An autoencoder, on the other hand, identifies a nonlinear function mapping the high-dimensional points to a low-dimensional latent representation without any metric, and while autoencoders are parametric in some sense, they are often trained with a large number of parameters, resulting in a high degree of flexibility in the final low-dimensional representation.}

%
%

{Our approach therefore is to use a MIXture of AutoEncoders (MIXAE), each of which should identify a non-linear mapping suitable for a particular cluster.} The autoencoders are trained simultaneously with a mixture assignment network via a composite objective function, thereby jointly motivating low reconstruction error (for learning each manifold) and cluster identification error. 
This kind of joint optimization has been shown to have good performance in other unsupervised architectures \cite{zheng2016variational} as well.
The main advantage in combining clustering with representation learning this way is that the two parts collaborate with each other to reduce the complexity and improve representation ability--the latent vector itself is a low-dimensional data representation that allows a much smaller classifier network for mixture assignment, and is itself learned to be well-separated  by clustering (see section \ref{sec:exp}). 


{Our contributions are: {\em{(i)}} a novel deep learning architecture for unsupervised clustering with mixture of autoencoders, {\em{(ii)}} a joint optimization framework for simultaneously learning a union of manifolds and clustering assignment, and {\em{(iii)}} 
state-of-the-art performance on established benchmark large-scale datasets. }

\section{Related Work} 
\label{sec:relwork}

{The most fundamental method for clustering is the K-means algorithm \cite{hartigan1979algorithm}, which assumes that the data are centered around some centroids, and seeks to find clusters that minimize the sum of the squares of the $\ell_2$ norm distances to the centroid within each cluster.  Instead of modeling each cluster with a single point (centroid), another approach called K-subspaces clustering assumes the dataset can be well approximated by a union of subspaces (linear manifolds); this field is well studied and a large body of work has been proposed \cite{vidal2011subspace,elhamifar2013sparse}. However, neither K-means nor K-subspaces clustering is designed to separate clusters that have {nonlinear and non-separable structure.}}

Nonlinear manifold clustering has been studied as a more promising generalization of linear models  and has an extensive literature~\cite{mordohai2005unsupervised,gionis2005dimension,von2007tutorial,wang2014riemannian,souvenir2005manifold,wang2010multi}. 
In particular, graph-based methods like spectral clustering \cite{shi2000normalized,ng2002spectral,von2007tutorial} and its variants \cite{yang2010image} are popular ways of handling nonlinearly separated clusters. However, these methods in general can be computationally intensive, and still have difficulties in separating clusters with intersecting regions.

Recent work \cite{tian2014learning} extends spectral clustering by replacing the eigenvector representation of data with the embeddings from a  deep autoencoder. 
Since training an autoencoder is linear in the number of samples $N$, finding the embeddings is much more scalable than traditional spectral clustering,  despite the difficulties in training autoencoders. {However, this approach requires a $N\times N$ normalized adjacency matrix as input, which is a heavy burden on both computation and memory for very large $N$.}

Mixture models are a computationally scalable probabilistic approach to clustering that also allows for overlapping clusters.  The most popular mixture model is the  Gaussian Mixture Model (GMM), which assumes that data are generated from a mixture of Gaussian distributions with unknown parameters, and the parameters are optimized by the Expectation Maximization (EM) algorithm. 
However, as with the K-means method, GMMs require strong assumptions on the distribution of the data, which are often not satisfied in practice.

A recent stream of work has focused on optimizing a clustering objective over the low-dimensional feature space of  an autoencoder \cite{xie2016unsupervised} or a variational autoencoder \cite{zheng2016variational,dilokthanakul2016deep}. 
Notably, the Deep Embedded Clustering (DEC) model \cite{xie2016unsupervised} 
iteratively minimizes the within-cluster KL-divergence and the reconstruction error.
The Variational Deep Embedding (VaDE)  \cite{zheng2016variational}  and Gaussian Mixture Variational Autoencoder (GMVAE) \cite{dilokthanakul2016deep} models extend the DEC approach by training variational autoencoders, iteratively learning clusters and feature representation distribution parameters.
In particular, \cite{zheng2016variational}  emphasizes a noticeable gain in training the autoencoder and the GMM components jointly rather than alternatively, which shares the same spirit of our joint representation and clustering framework. 
A weakness in these models is that  they require careful initialization of model parameters, and often exhibit separation of clusters before actual training even begins. {In contrast, the proposed MIXAE model can be trained from scratch. }

Similarly, the DLGMM model \cite{nalisnick2016approximate} and CVAE model  \cite{shu2016stochastic} also combine variational autoencoders with GMM for clustering, but are primarily used for different applications.
Adversarial autoencoders \cite{makhzani2015adversarial} are another popular extension, and both are also popular for semi-supervised learning \cite{makhzani2015adversarial,abbasnejad2016infinite}.
However, adversarial models are reputably difficult to train.

\section{Clustering with Mixture of Autoencoders} 
\label{sec:model}
\begin{figure*}[tb]
	\centering
	\includegraphics[width=5.5in, height=2.4in]{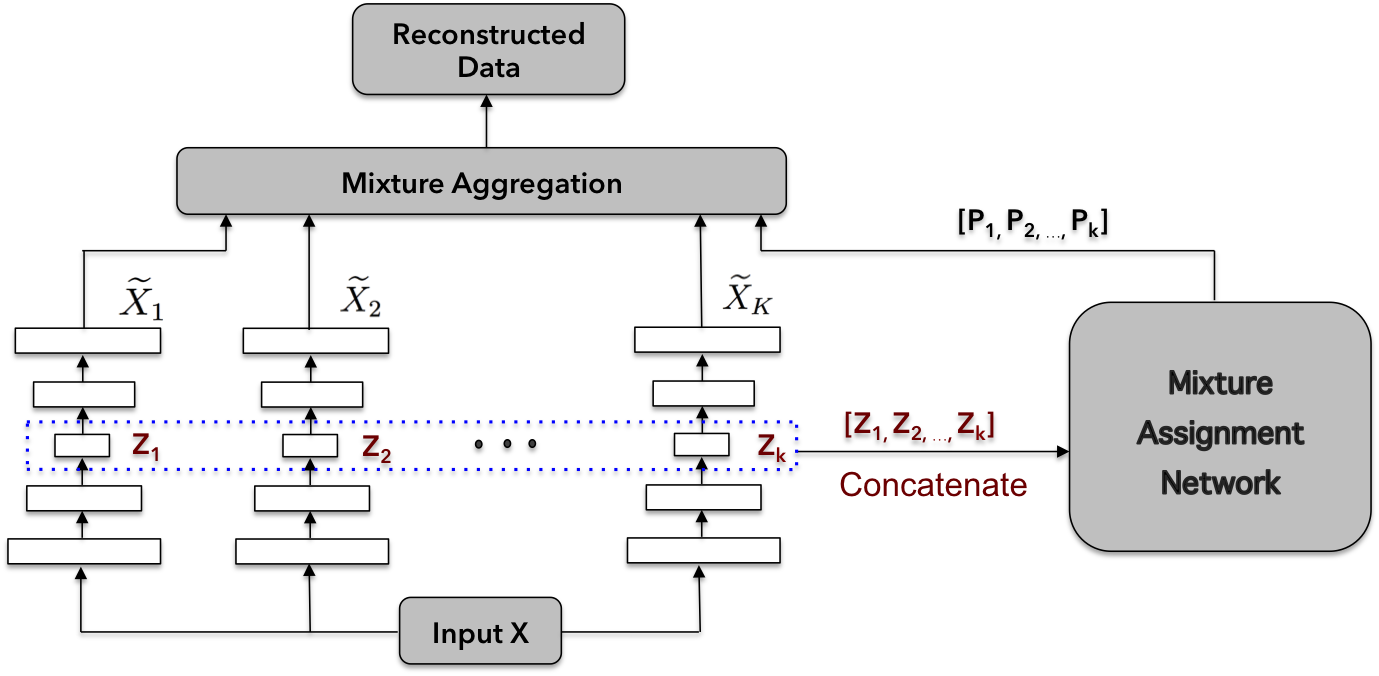}
	\caption{\textbf{Overall architecture of the MIXAE model}. The MIXAE architecture contains several parts: (a) a collection of $K$ autoencoders, each of them seeking to learn the underlying manifold of one data cluster; (b) for each input data, the mixture assignment network takes the concatenated latent features as input and outputs soft clustering assignments; (c) the mixture aggregation which is done via the weighted reconstruction error together with proper regularizations on $\mathbf{p}^{(i)} = [p_1^{(i)}, \dots, p_K^{(i)}]$.}
	\label{fig:mixae_model}
\end{figure*}

We now describe our MIXture of AutoEncoders (MIXAE) model in detail, giving the intuition behind our customized architecture and specialized objective function.

\subsection{Autoencoders} 
\label{sub:autoencoder}
An autoencoder is a common neural network architecture used for unsupervised representation learning. An autoencoder consists of an encoder $(\mathcal{E})$ and a decoder $(\mathcal{D})$. 
Given the input data $x \in \R^{n}$, the encoder first maps $x$ to its latent representation $z = \mathcal{E}(x) \in \R^{d}$, where typically $d < n$. The decoder then maps $z$ to a reconstruction $\tilde x = \mathcal{D}\left(z\right) \in \R^{n}$, with reconstruction error measuring the deviation between $x$ and $\tilde x$. 
The parameters of the network are updated via backpropagation with the target of minimizing the reconstruction error. By restricting the latent space to lower dimensionality than the input space ($d < n$) the trained autoencoder parametrizes a low-dimensional nonlinear manifold that captures the data structure.

\subsection{Mixture of Autoencoders (MIXAE) Model} 
Our goal is to cluster a collection of data points $\left\{x^{(i)}\right\}_{i=1}^N \in \R^{n}$ into $K$ clusters, under the assumption that data from each cluster is sampled from a different low-dimensional manifold.  
A natural choice is to use a separate autoencoder to model each data cluster, and thereby the entire dataset as a collection of autoencoders.   
The cluster assignment is performed with an additional neural network, which infers
the cluster labels of each data sample based on the autoencoders' latent features.  Specifically, for each data sample $x_i \in \mathbb{R}^{n}$, this mixture assignment network takes the concatenation of the latent representations of each autoencoder \[
\mathbf z^{(i)} = (z^{(i)}_1,\ldots,z^{(i)}_K) = (\mathcal{E}_1(x^{(i)})),\ldots,\mathcal{E}_K(x^{(i)}))\in \mathbb R^{dK} 
\] as input, and outputs a probabilistic vector $\mathbf{p}^{(i)} = [p_1^{(i)}, \dots, p_K^{(i)}]$ that infers the distribution of $x_i$ over clusters, \ie for $k = 1, \dots, K$, 
\begin{align}
{p}_k^{(i)} &= \text{Pr}\left[x_i \in \text{cluster $k$} \big \lvert \mathbf z^{(i)}\right]. 
\label{eq: cluster_infer}
\end{align} 
To jointly optimize the $K$ autoencoders and the mixture assignment network, we use a composite objective function consisting of three important components.

\paragraph{Weighted reconstruction error} The mixture aggregation is done in the weighted reconstruction error  term
\begin{multline}
\label{eq:reconstruct_error}
\textbf{Reconstr}\left(x^{(i)}, \{\widetilde x^{(i)}_k\}_{k=1}^{K}, \mathbf p^{(i)}\right) \\
=\sum_{k = 1}^{K} {p}^{(i)}_k\mathcal{L}\left(x^{(i)}, \widetilde x_k^{(i)}\right)
\end{multline}
where $x^{(i)}$ is the $i^{th}$ data sample, $\widetilde x^{(i)}_k$ is the reconstructed output of autoencoder $k$ for sample $i$, $\mathcal{L}(x, \widetilde x)$ is the reconstruction error, and $ p_k^{(i)}$ are the soft probabilities from the mixture assignment network for sample $i$, calculated via \eqref{eq: cluster_infer}.
Typical choices for $\mathcal{L}$ are squared errors and KL-divergence. 
Intuitively, (\ref{eq:reconstruct_error}) will achieve its minimum when $\mathbf{p}^{(i)}$'s are one-hot vectors and select the autoencoder with minimum reconstruction error for that sample.

\paragraph{Sample-wise entropy}
To motivate sparse mixture assignment probabilities (so that each data sample ultimately receives one dominant label assignment)
we add a sample entropy deterrent:
\begin{align}
\textbf{Entr}({\mathbf p^{(i)}}) := - \sum_{k = 1}^{K} p^{(i)}_k \log (p_k^{(i)}) .
\label{eq:sample_entropy}
\end{align}  
Specifically, (\ref{eq:sample_entropy}) achieves its minimum $0$ only if $\mathbf p^{(i)}$ is an one-hot vector, specifying a deterministic distribution. 
We refer to this as the \textit{sample-wise entropy}.
\vspace{-0.2cm}
\paragraph{Batch-wise entropy:}  
One trivial minimizer of the sample-wise entropy is for the mixture assignment neural network to output a constant one-hot vector $\mathbf {p}^{(i)}$ for all input data, \ie  selecting a single autoencoder for all of the data. To avoid this local minima, we motivate equal usage of all autoencoders via a \emph{batch-wise entropy} term
\begin{align}
\textbf{Entr}({\bar {\mathbf{p}}^{(i)}}) &:= - \sum_{k = 1}^{K} \bar{p}^{(i)}_k \log (\bar{p}^{(i)}_k)\;, \nn \\
&\text{where}~~{\bar {\mathbf{p}}^{(i)}} := \frac{1}{\cal{B}}\sum_{i=1}^{{\cal{B}}} \mathbf {p}^{(i)} \;. \label{eq:batch_entropy}   
\end{align}
Here, $\mathcal{B}$ is the minibatch size and $\mathbf {\bar p}$ is the average soft cluster assignment over an entire minibatch. 
If (\ref{eq:batch_entropy}) reaches a maximum value of $\log(K)$, then $\mathbf {\bar{p}} = \frac{1}{K}\textbf{1}$, \ie each cluster is selected with uniform probability. 
This is a valid assumption for a large enough minibatch, randomly selected over balanced data.

\paragraph{Objective function} 
Let $\theta=(\theta_1,\ldots,\theta_K,\theta_\text{MAN})$ be the parameters of the autoencoders and mixture assignment network. We minimize the composite cost function 
\begin{multline}
\mathcal{J}(\theta) = \frac{1}{|\mathcal{B}|}\sum_{i\in\mathcal{B}} \left(\textbf{Reconstr}\left(x^{(i)}, \{\widetilde x^{(i)}\}_{k=1}^{K}, \mathbf p^{(i)}\right) \right. \\
\left . + \alpha \textbf{Entr}(\mathbf{p}^{(i)})\vphantom{\sum_{\mathcal B}} \right) - \beta\textbf{Entr}(\mathbf {\bar{p}}).
\label{eq:total_loss}
\end{multline}

An important consideration is the choice of $\alpha$ and $\beta$, which can significantly affect the final clustering quality. {We adjust these parameters dynamically during the training process.} Intuitively, initially we should prioritize batch-wise entropy and sample-wise entropy in order to encourage equal use of autoencoders while avoiding the case where all autoencoders are equally optimized for each input, \ie the probabilistic vector characterizes a uniform distribution for each input. 
Asymptotically, we should prioritize minimizing the reconstruction error {to promote better learning of the manifolds for each cluster, and minimizing sample-wise entropy to ensure assignment of every data sample to only one autoencoder.} A simple and effective scheme is to use comparatively larger $\alpha$ and $\beta$ at the beginning of the training process, while adjusting $\alpha$ and $\beta$ at each epoch such that the three terms in the objective function are approximately equal as the training process goes on. Empirically, this produces better results than static choices of $\alpha$ and $\beta$.

\section{Experimental Results} 
\label{sec:exp}
We evaluate our MIXAE on three datasets representing different applications: images, texts, and sensor outputs.

\paragraph{MNIST}{
	The MNIST \cite{lecun1998gradient} dataset contains 70000  $28\times 28$ pixel images of handwritten digits (0, 1, \dots, 9), each cropped and centered.
}

\paragraph{Reuters}{The original Reuters dataset contains about 810000 English news stories labeled by a category tree. Following \cite{xie2016unsupervised}, we choose four root categories: corporate/industrial, government/social, markets, and economics as labels, and remove all of the documents that are labeled by multiple root categories, which results in a dataset with 685071 documents.  We then compute the tf-idf features  on the 2000 most frequent words.}

\paragraph{HHAR}{
	The Heterogeneity Human Activity Recognition (HHAR, \cite{stisen2015smart}) dataset contains  10299 samples of smartphone and smartwatch sensor time series feeds, each of length 561. There are 6 categories of human activities: walking, walking upstairs, walking downstairs, sitting, standing, and laying.
	}

A summary of the dataset statistics is also provided in Table \ref{tab:dataset}.
\begin{figure*}[tb]
	\begin{center}
		\includegraphics[width=6.5in]{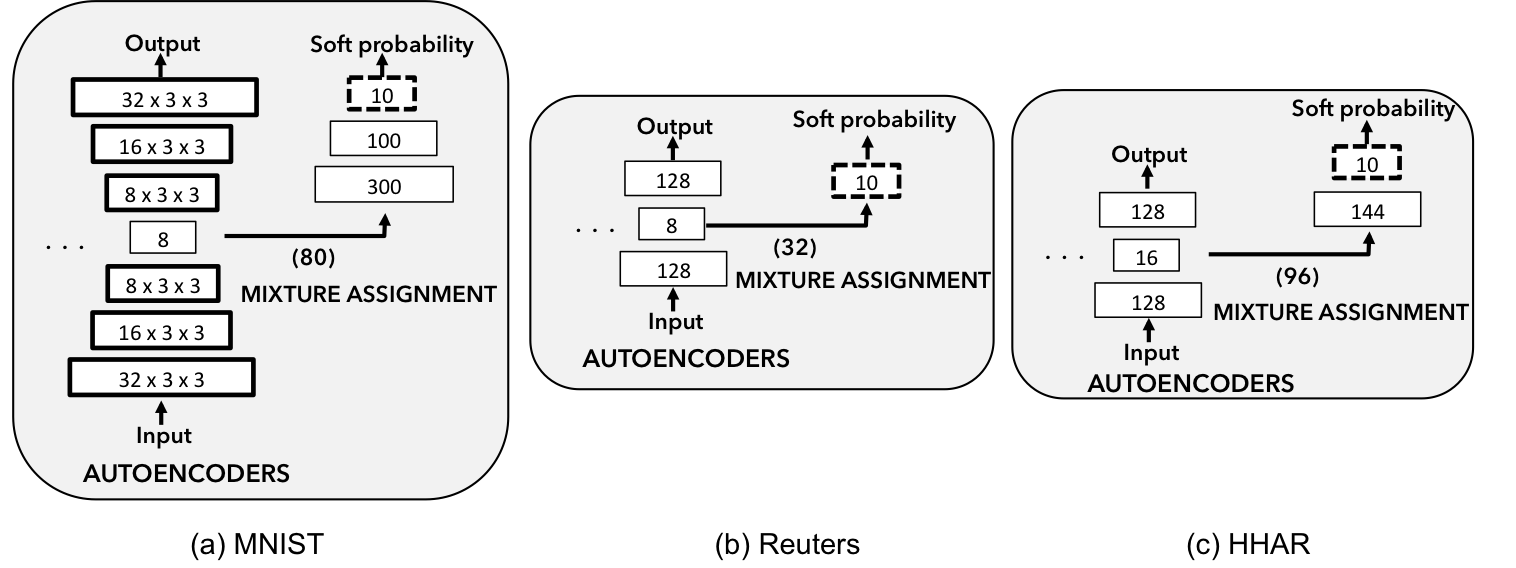}
	\end{center}
	\caption{\textbf{Network Architecture.} Autoencoder and mixture assignment networks for (a) MNIST, (b) Reuters, and (c) HHAR experiments. Thin solid, thick solid, and dashed lines  show the output of fully-connected, CNN, and softmax layers respectively.}
	\label{fig:network_architecture}
\end{figure*}

\paragraph{Network architecture}{
	We use different autoencoders and mixture assignment network sizes for different datasets, summarized in Figure \ref{fig:network_architecture}.  We use convolutional autoencoders for MNIST and  fully connected autoencoders for the other (non-image) datasets.  For each dataset, we train MIXAE with ADAM \cite{kingma2014adam} acceleration, using Tensorflow. We use a decaying learning rate, initialized at $0.001$ and reduced by a factor $0.9$ every $10$ epochs.}
\begin{table}[H]
	\caption{\textbf{Datasets.} $N=$ \# samples. $K=$ \# clusters. The last column shows class balance by giving the percent of data in the largest class (LC) / smallest class (SC).}
	\label{tab:dataset}
	\centering
	\begin{tabular}{l|c|c|c|c}
		\hline
		
		\hline
		\textbf{Dataset} &  \text{Dim} &\text{$n$} & $K$ &LC /SC\\
		\hline
		MNIST	 &784 & 70000 & 10 & 11\% / 9\% \\
		Reuters &2000 &685071 & 4& 41\% / 9\%\\
		HHAR & 561 & 10299 & 6& 19\% / 14\%\\
		\hline
	\end{tabular}
\end{table}

\paragraph{Evaluation metric} 
Following the work of DEC, the clustering accuracy of all algorithms is measured by the \textit{unsupervised clustering accuracy (ACC)}: 
\begin{equation}
\text{ACC} =\max_{m \in \mathcal{M}}  \frac{1}{N}\sum_{i=1}^{N} \textbf{1}\left\{l_i = m(c_i)\right\}
\label{eq:accur_metric}
\end{equation}
where $l_i$ is the ground-truth label, $c_i$ is the cluster assignment produced by the mixture assignment network, \ie 
\[
c_i = \arg\max_k  p_k^{(i)}
\]
and $m\in \mathcal{M}$ are all possible one-to-one mappings between clusters and labels. 
This is equivalent to the \emph{cluster purity} and is a common metric in clustering (see also \cite{xie2016unsupervised}). 
Finding the optimal mapping can be done effectively using the Hungarian algorithm \cite{munkres1957algorithms}.

\subsection{Clustering results}	
An overall comparison of each clustering method is given in Table \ref{tab:cluster_accur}.
As we can see, the  deep learning models (DEC, VaDE and MIXAE) all perform much better than traditional machine learning methods (K-means and GMM). 
This suggests that using autoencoders to extract the latent features of the data and then clustering on these latent features is  advantageous for these challenging datasets. 

Comparing the  deep learning models, we see that MIXAE outperforms DEC, a single manifold model, suggesting the advantage of a mixture of $K$ manifolds in cluster representability.
Additionally, note that both DEC and VaDE use stacked autoencoders  to pretrain their models, which can introduce significant computational overhead, especially if the autoencoders are deep.  In contrast, MIXAE trains from a random initialization.

At the same time,  though MIXAE achieves the same or slightly better performance against VaDE on Reuters and HHAR,  VaDE outperforms MIXAE and DEC on MNIST.  
In general it has been observed that variational autoencoders have better representability than deterministic autoencoders (\eg~ \cite{kingma2013auto}).
This suggests that  leveraging a mixture of variational autoencoders may  further improve our model's performance and is an interesting direction for future work.

\begin{table}
	\caption{\textbf{Clustering accuracy.} Comparison of unsupervised clustering accuracy (ACC) on different datasets.  }
	\label{tab:cluster_accur}
	\centering
	\begin{tabular}{l|cccc}
		\hline
		\textbf{Method} & \textbf{MNIST} & \textbf{Reuters} & \textbf{HHAR} & \textbf{Pretrain?}\\
		\hline
		K-means&  53.5\%& 53.3\% & 60.1\%& No\\
		GMM&   53.7\% &55.8\% & 60.3\% & No \\
		DEC \cite{xie2016unsupervised}&  84.3\% & 75.6\%& 79.9\% &Yes\\  
		VaDE \cite{zheng2016variational}& \textbf{94.5}\% & \textbf{79.4}\%& 84.5\%& Yes\\
		\hline
		MIXAE&    85.6\% & \textbf{79.4}\% &\textbf{87.8}\%\  & No\\
		\hline
	\end{tabular}
\end{table}
	
\begin{figure}[tb]
	\centering
	\includegraphics[scale=0.42]{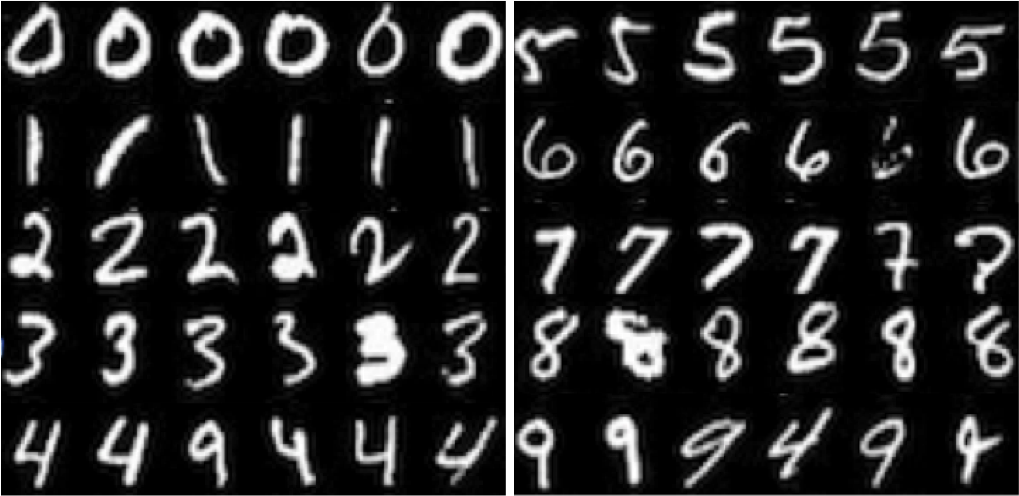}
	\caption{\textbf{Sample output.} Visualization of the clustering results on MNIST for $K=10$. (Each row within each subfigure is a cluster.)}
	\label{fig:visu_clusters}
\end{figure}
	
\begin{figure*}[tb]
	\centering
	\includegraphics[scale=0.36, trim={4.5cm 2.0cm 0.0cm 0cm},clip]{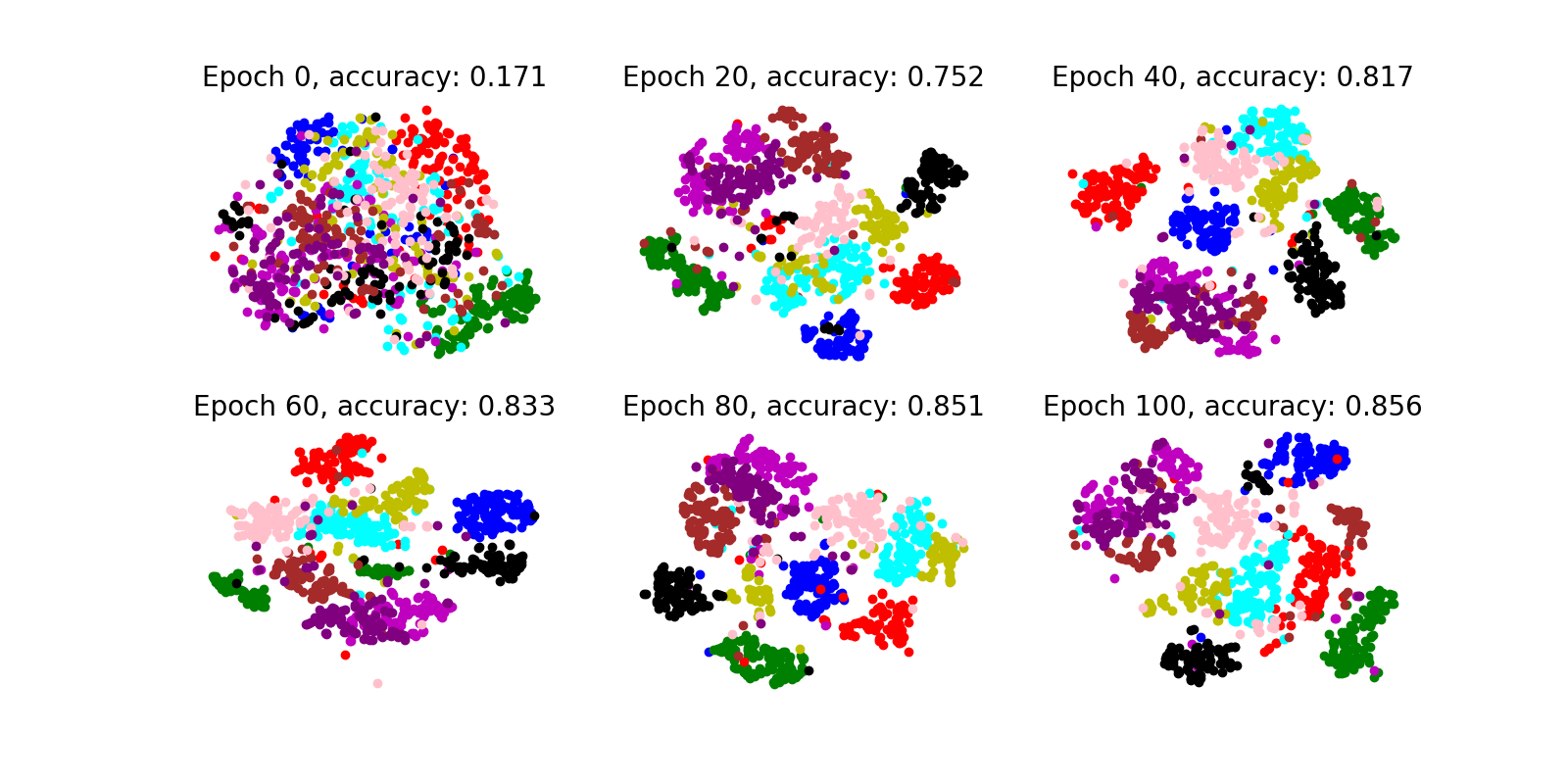}
	\caption{\textbf{Latent space visualization.} t-SNE projection of the 80-dim concatenated latent vectors from the MNIST experiment, projected to a 2-dimensional space  \cite{maaten2008visualizing}. }
	\label{fig:latent_Z}
\end{figure*}	
Figure \ref{fig:visu_clusters} shows some samples grouped by cluster label. We see that MIXAE clusters well a variety of writing styles. There are some mistakes, but are consistent with frequently observed mistakes in supervised classification (\eg 4 and 9 confusion).

\subsection{Latent space separability}
 Fig \ref{fig:latent_Z} shows the t-SNE projection of the $d K$-dimensional concatenated latent vectors to 2-D space. Specifically, we see that as training progresses, the latent feature clusters become more and more separated, suggesting that the overall architecture motivates finding representations with better clustering performance.

\subsection{Effect of balanced data} 
\begin{figure}[tb]
	\centering
	\includegraphics[scale=0.31, trim={3.0cm 2.0cm 0.0cm 0cm},clip]{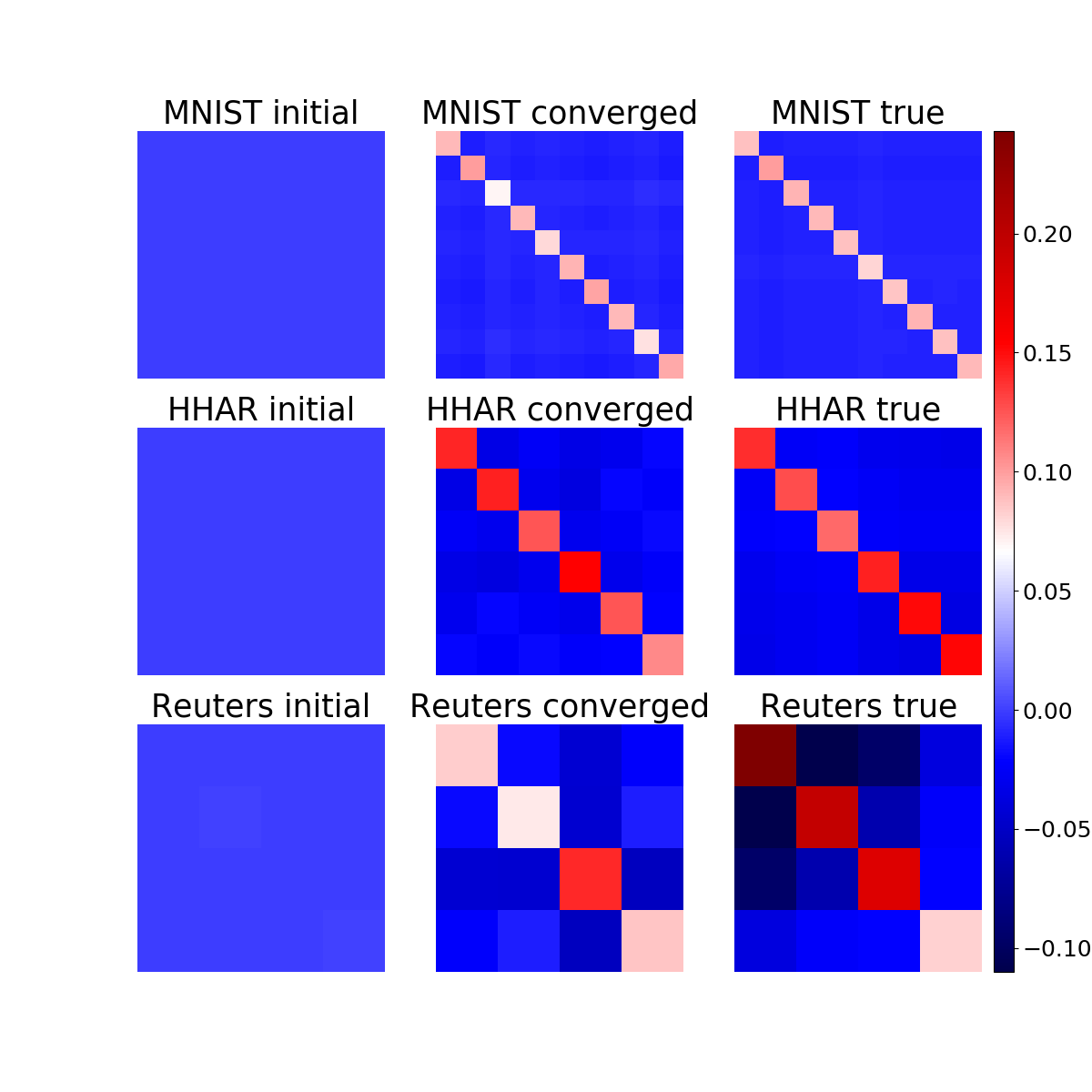}
	\caption{\textbf{Clustering covariance.}  Covariance matrices  $X = \frac{1}{N}(\sum_i\mathbf{p}^{(i)} (\mathbf p^{(i)})^T) - \mathbf{\bar p} \mathbf {\bar p}^T$, where $\mathbf{p}^{(i)}\in [0,1]^K$ is the output of the mixture assignment network at sample $i$, and   $\mathbf{\bar p} = \frac{1}{N}\sum_i \mathbf{p}^{(i)}$.
	The true covariance matrix is computed similarly, but replacing $\mathbf p^{(i)}$ with the one-hot representation of the true label of sample $i$.
}
	\label{fig:qy_visu}
\end{figure}

As we can see in Table \ref{tab:cluster_accur}, all methods have significantly lower performance on Reuters (an unbalanced dataset) than MNIST and HHAR (balanced datasets).   We investigate  the effect of balanced data on MIXAE in Table \ref{tab:batch_entropy} and Figure \ref{fig:qy_visu}. 
\begin{table}
	\caption{\textbf{Data imbalance.} Left: the expected batch entropy assuming uniform clusters (max value)  and given cluster imbalance. Right: the actual batch and sample entropy values for each dataset. Max BE (batch entropy) $= \log(K)$. Expected BE $\sum_k p_k\log(p_k)$, where $p_k =$ \# samples with label $k$ / \# samples.  Actual BE and SE (sample entropy) are converged values.}
	\label{tab:batch_entropy}
	\centering
	\begin{tabular}{l|ccc|cc}
		\hline
		&\multicolumn{3}{c|}{Ground Truth}	&\multicolumn{2}{c}{Actual}\\\hline
		 \textbf{Dataset} & $\log(K)$ &  BE &SE &  BE & SE \\
		 \hline
		 \textbf{MNIST} & 2.31&2.30 &0&2.28 &0.026\\
		  \textbf{Reuters} &1.39& 1.26 &0& 1.38&0.598\\
		   \textbf{HHAR} & 1.79&1.79 &0& 1.77&0.054\\
		\hline
	\end{tabular}
\end{table}

In Table \ref{tab:batch_entropy}, for each dataset, we record the values of batch-wise entropy (BE) and sample-wise entropy (SE) over the entire dataset after training, and we compare them with the ground truth entropies of the true labels.  The batch entropy regularization \eqref{eq:batch_entropy} forces the final actual batch-wise entropy to be very close to the maximal value of $\log(K)$ for all of the three datasets. 

\begin{figure*}[htpb]
\centering
\includegraphics[scale=0.42,trim={1.5cm 0.0cm 0.0cm 0cm},clip]{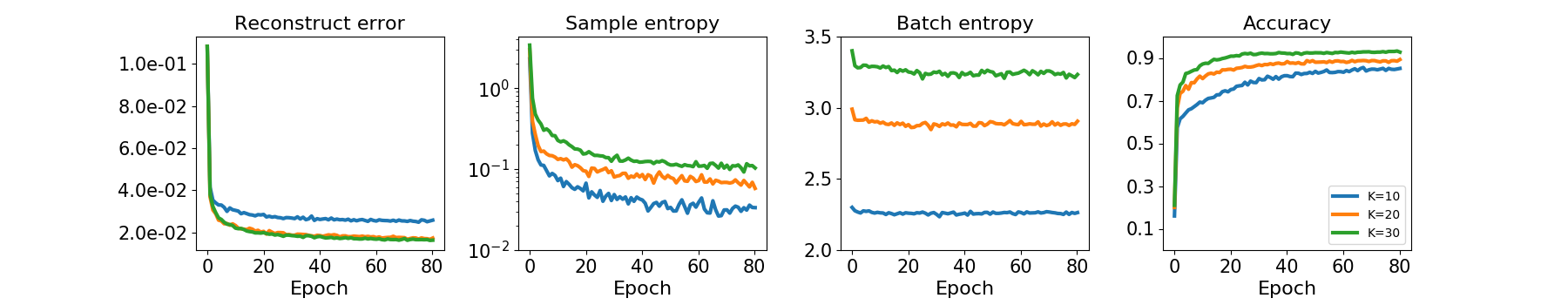}
\caption{\textbf{Evolution for varying $K$.} Key components of the  objective function  \eqref{eq:total_loss} during the training over the MNIST dataset, for $K = 10$, $20$, and $30$. }
\label{fig:Obj_var_K}
\end{figure*}

In other words, our batch entropy encourages the final cluster assignments to be uniform, which is valid for balanced datasets (MNIST and HHAR) but  biased for unbalanced datasets (Reuters). 		
Specifically, in Fig.~\ref{fig:qy_visu}, the sample covariance matrix of the true labels of Reuters has one dominant diagonal value, but the converged sample covariance matrix diagonal is much more even, suggesting that samples that should have gone to a dominant cluster are evenly (incorrectly) distributed to the other clusters.
 
Additionally, note that the converged sample-wise entropy (actual SE value) for Reuters is far from 0 (Table \ref{tab:batch_entropy}), suggesting that even after convergence, there is considerable ambiguity in cluster assignment.

\subsection{Varying K} 
We also explore the clustering performance of MIXAE with more autoencoders than natural clusters; \ie for MNIST, $K = 20$ and $K = 30$. 
In Figure \ref{fig:Obj_var_K}, we plot the evolution of the three components of our objective function  \eqref{eq:total_loss}, as well as the final cluster purity.	
This purity is defined as the percentage of ``correct" labels, where the ``correct" label for a cluster is defined as the majority of the  true labels for that cluster.

\begin{figure}[tb]
	\begin{center}
		\begin{subfigure}{0.9\columnwidth}
			\includegraphics[scale=0.32,trim={4.0cm 0.0cm 0.0cm 0cm},clip]{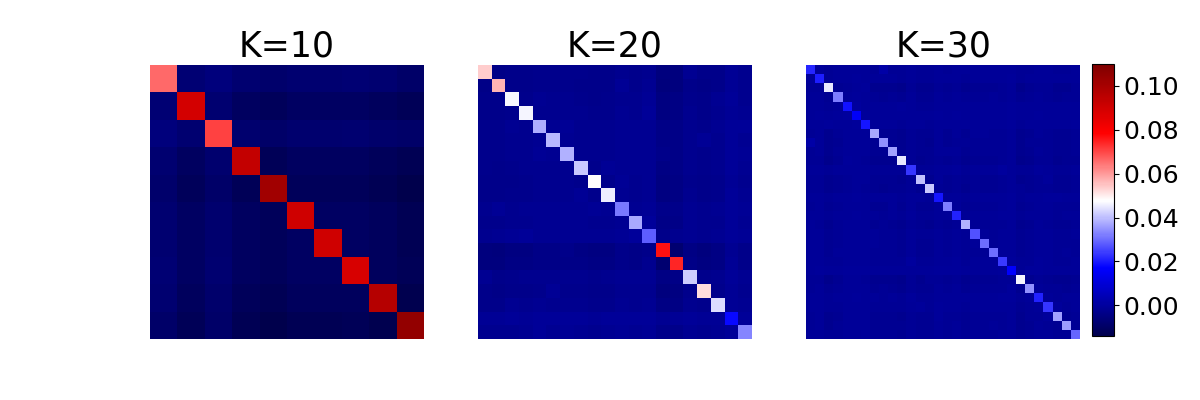}
			\caption{Covariance matrix of the estimated $\mathbf{p}^{(i)}$ vector.}
			\label{fig:figure1}
		\end{subfigure}
		\begin{subfigure}{0.9\columnwidth}
			\includegraphics[scale=0.6,trim={2.0cm 2.2cm 1.5cm 2.0cm},clip]{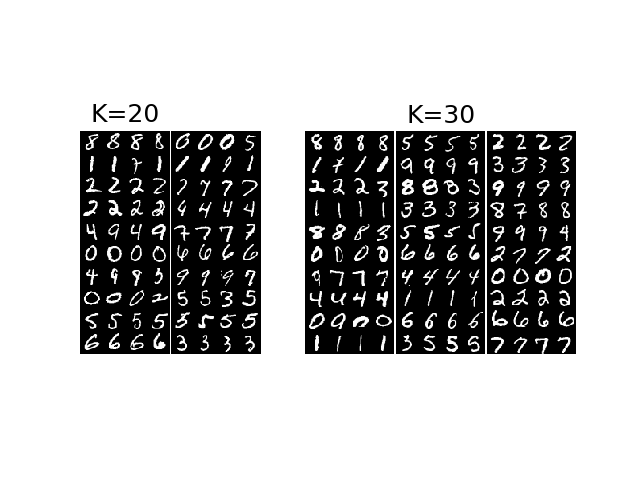}
			\caption{Sample outputs (4 per cluster).}
			\label{fig:figure2}
		\end{subfigure}
	\end{center}
	\caption{\textbf{Variant K.} Visualization of the clustering results of MNIST with $K = 20$ and $K = 30$.}
	\label{fig:visu_var_K}
\end{figure}

As we can see in Figure \ref{fig:visu_var_K}, the clustering accuracy for larger $K$ converges to higher values. One possible explanation is that with larger $K$, the final clusters split each digit group into more clusters, and this reduces the overlap in underlying manifolds corresponding to different digit groups. 
On the other hand, the sample-wise entropy no longer converges to 0, and the final probabalistic vectors are observed to have 2 or 3 significant nonzeros instead of only one; this suggests that the learned manifolds corresponding with each digit group may have certain overlap.

Figure \ref{fig:figure1} shows again the covariance matrices for MNIST, $K = 10, 20$, and $30$. 
Interestingly, here the final covariance diagonals are extremely uneven, suggesting that final cluster assignments are more and more unbalanced as we increase $K$. 
Since intuitively  each digit group may have different magnitudes of variance in writing styles, this result is consistent with what we may expect.	
Figure \ref{fig:figure2} shows digit examples, sorted according to the finalized cluster assignments. Here, we see the emergence of ``stylistic" clusters, with straight vs slanted 1's, thin vs round 0's, etc. 

\section{Conclusions and Future Work} 
\label{sec:conclude}
In this paper, we introduce the MIXAE architecture that uses a combination of small autoencoders and a cluster assignment network to intelligently cluster unlabeled data. This is based on the assumption that data from each cluster is generated from a separate low-dimensional manifold, and thus the aggregate data is modeled as a mixture of manifolds. 
Using this model, we produce improved performance over deterministic deep clustering models on established datasets.

There are several interesting extensions to pursue. First, though we have improved performance on the unbalanced dataset over DEC, we still find Reuters a challenging dataset due to its imbalanced distribution over natural clusters.
One potential improvement is to replace the batch entropy regularization with cross-entropy regularization, using knowledge about cluster sizes. However, knowing the sizes of clusters is not a realistic assumption in online machine learning. Additionally this would force each autoencoder to take a pre-assigned cluster identity, which might negatively affect the training. 

Another important extension is in the direction of variational and adversarial autoencoders. We have seen that in single autoencoder models,  VaDE outperforms DEC, which they also attribute to a KL penalty term for encouraging cluster separation. This extension can be done in our model to encourage separation in the latent representation variables.

Our model also has an interesting interpretation to  dictionary learning, where a small set of basis vectors characterizes a structured high dimensional dataset. Specifically, we can consider the manifolds learned by the autoencoders as ``codewords" and the sample entropy applied to the mixture assignment as the sparse regularization. An interesting extension is to apply this model to multilabel clustering, to see if each autoencoder can learn distinctive atomic features of each datapoint--for example, the components of an image, or voice signal.

{\small
\bibliographystyle{ieee}
\bibliography{egbib}
}

\end{document}